\documentclass[runningheads]{llncs}
\usepackage[compatibility=false]{caption}
\usepackage[T1]{fontenc}
\usepackage{subcaption}
\usepackage{graphicx}
\usepackage{adjustbox}
\usepackage{tabularx}
\usepackage[table]{xcolor}
\usepackage{xspace}
\usepackage{amsfonts}
\usepackage{amsmath}
\usepackage{amssymb}
\usepackage{bm}
\newcommand{\green}[1]{\textcolor{green!60!black}{#1}}
\newcommand{\blue}[1]{\textcolor{blue!80!black}{#1}}
\newcommand{\red}[1]{\textcolor{red!75!black}{#1}}
\newcommand{\ie}{\emph{i.e.,}\xspace}
\newcommand{\eg}{\emph{e.g.,}\xspace}
\newcommand{\review}[1]{\textcolor{black}{#1}}
\newcommand{\reviewData}[1]{\textcolor{black}{#1}}

\begin{document}
%
\title{FLIM Networks with Bag of Feature Points}
\titlerunning{FLIM Networks with Bag of Feature Points}

\author{João Deltregia Martinelli\inst{1}\orcidID{0000-0002-1249-5317} \and
Marcelo Luis Rodrigues Filho \inst{1}\orcidID{0009-0002-3848-4795} \and
Felipe Crispim da Rocha Salvagnini \inst{1}\orcidID{0000-0002-7896-0058} \and
Gilson Junior Soares \inst{1}\orcidID{0000-0001-8805-5644} \and
Jefersson A. dos Santos\inst{2}\orcidID{0000-0002-8889-1586} \and 
Alexandre X. Falcão\inst{1}\orcidID{0000-0002-2914-5380}}
\authorrunning{Martinelli et al.}
%
\institute{Institute of Computing\\ UNICAMP\\ Campinas, Brazil
\and
School of Computer Science\\ University of Sheffield\\ Sheffield, United Kingdom \\
}

\maketitle              
\begin{abstract}
Convolutional networks require extensive image annotation, which can be costly and time-consuming. \textit{Feature Learning from Image Markers} (FLIM) tackles this challenge by estimating encoder filters (\textit{i.e.}, kernel weights) from user-drawn markers on discriminative regions of a few representative images without traditional optimization. Such an encoder combined with an \textit{adaptive decoder} comprises a FLIM network fully trained without backpropagation. Prior research has demonstrated their effectiveness in Salient Object Detection (SOD), being significantly lighter than existing lightweight models. This study revisits FLIM SOD and introduces FLIM-\textit{Bag of Feature Points} (FLIM-BoFP), a considerably faster filter estimation method. The previous approach, FLIM-Cluster, derives filters through patch clustering at each encoder's block, leading to computational overhead and reduced control over filter locations. FLIM-BoFP streamlines this process by performing a single clustering at the input block, creating a bag of feature points, and defining filters directly from mapped feature points across all blocks. The paper evaluates the benefits in efficiency, effectiveness, and generalization of FLIM-BoFP compared to FLIM-Cluster and other state-of-the-art baselines for parasite detection in optical microscopy images.

\end{abstract}
\section{Introduction}
\setcounter{footnote}{0}
\label{sec:intro}
Convolutional Neural Networks (CNNs) are essential for image analysis but increasingly rely on large annotated datasets. In domains requiring expert annotation, time constraints make manual labeling costly, tedious, and prone to errors, impacting CNN performance. Training complex CNNs from scratch also demands substantial computational resources, making it impractical in constrained environments. \review{Moreover, even running off-the-shelf CNNs in inference mode under restrictive scenarios (\eg mobile devices or low-powered computers in clinical environments of developing countries) might not be possible}.

\begin{figure}
    \centering
    \includegraphics[width=1\linewidth]{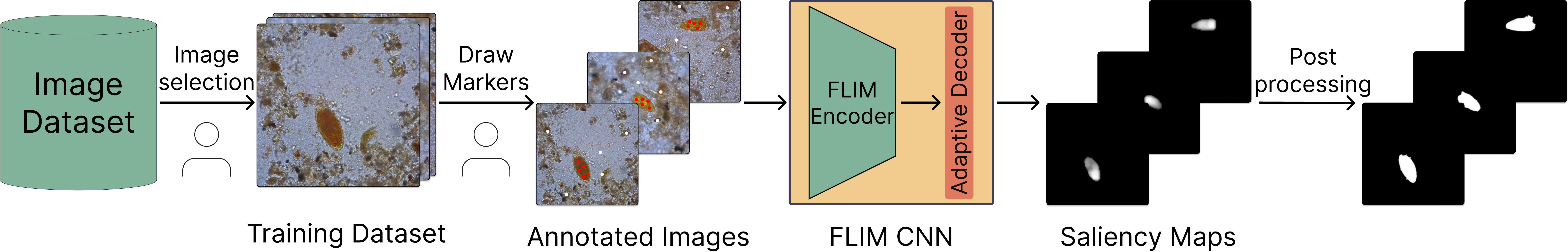}
    \vspace{0.1em}
    \caption{Training pipeline of a FLIM SOD Network from user-drawn markers on three representative images. In this example, the markers are red disks on the object (\textit{Schistosoma Mansoni} eggs) and white disks on the background (impurities and empty region). }
    \label{fig:flim-pipeline}
\end{figure}

\textit{Feature Learning from Image Markers} (FLIM) is a methodology for creating convolutional encoders without traditional optimization \review{(\ie backpropagation)}, deriving filters from patches in discriminative regions marked by users on a few representative images ~\cite{FLIM-og}. This approach has led to competitive lightweight models in various applications~\cite{FLIM-delineation,decoder-1}. A FLIM encoder, combined with an \textit{adaptive decoder}, enables flyweight networks ~\cite{decoder-1}—tens to hundreds of times lighter than existing lightweight models—for \textit{Salient Object Detection} (SOD). Since the adaptive decoder estimates its parameters for each image using a heuristic function, the entire CNN trains without backpropagation from minimal annotation requirements (Figure ~\ref{fig:flim-pipeline}).

 \review{\textit{FLIM methodology learns convolutional filters from image patches through clustering.} For each marker, patches are extracted from regions centered at marker pixels, normalized by z-score, and grouped, such that the cluster centers derive the filters of the current block.  For the first block, patches come from the input image, while subsequent blocks extract patches from feature maps generated by the previous convolutional block. The markers are mapped to the input of each block for patch extraction, normalization, and clustering.}

Cluster centers represent key visual patterns, with filters activating similar patterns at the block’s output. Some approaches refine these clusters using labeled markers to assign filters per class~\cite{FLIM-og}. While effective, per-block clustering increases computational overhead and limits filter placement control, affecting efficiency and explainability as network depth grows. 

\begin{figure}
    \centering
    \begin{subfigure}{0.19\textwidth}
        \centering
        \includegraphics[width=\textwidth]{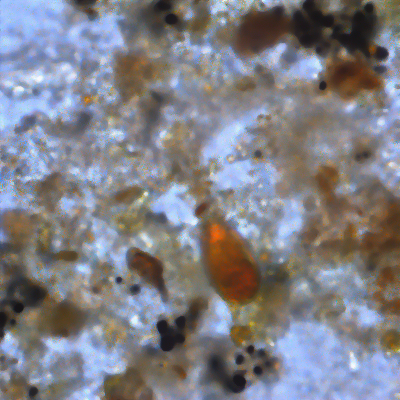}
        \caption{}
        \label{original}
    \end{subfigure}
    \begin{subfigure}{0.19\textwidth}
        \centering
        \includegraphics[width=\textwidth]{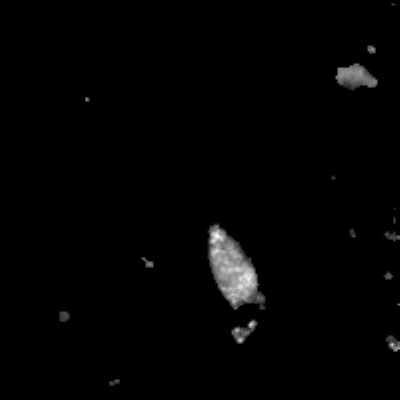}
        \caption{}
        \label{bofp1}
    \end{subfigure}
    \begin{subfigure}{0.19\textwidth}
        \centering
        \includegraphics[width=\textwidth]{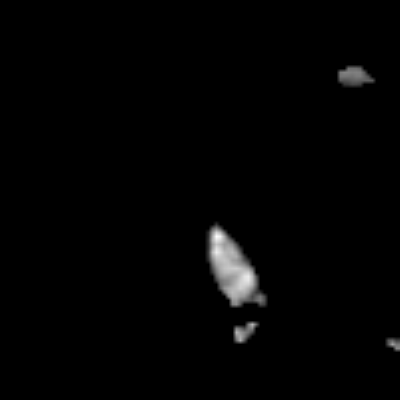}
        \caption{}
        \label{bofp2}
    \end{subfigure}
    \begin{subfigure}{0.19\textwidth}
        \centering
        \includegraphics[width=\textwidth]{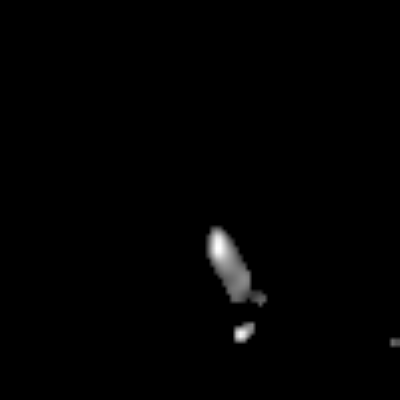}
        \caption{}
        \label{bofp3}
    \end{subfigure}
    \begin{subfigure}{0.19\textwidth}
        \centering
        \includegraphics[width=\textwidth]{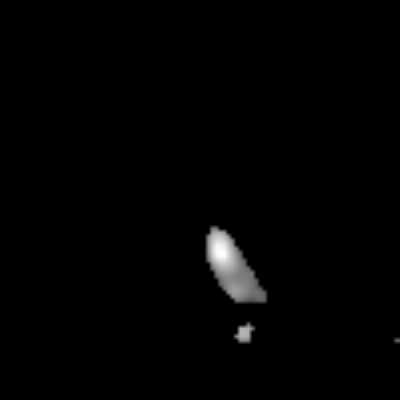}
        \caption{}
        \label{bofp4}
    \end{subfigure}
    
    \begin{subfigure}{0.19\textwidth}
        \centering
        \includegraphics[width=\textwidth]{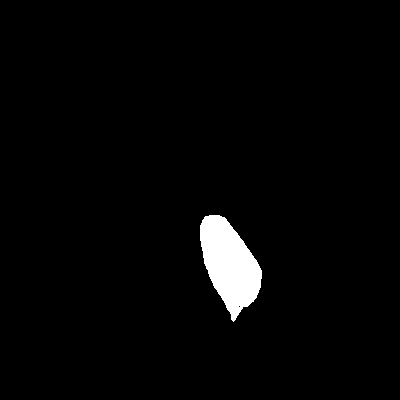}
        \caption{}
        \label{truelabel}
    \end{subfigure}
    \begin{subfigure}{0.19\textwidth}
        \centering
        \includegraphics[width=\textwidth]{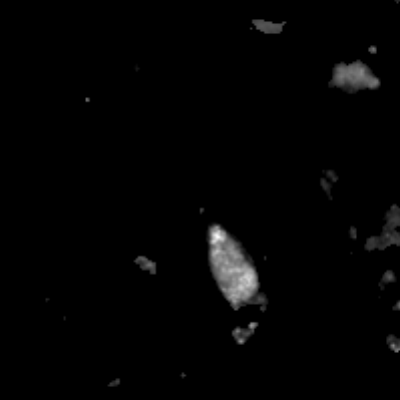}
        \caption{}
        \label{clust1}
    \end{subfigure}
    \begin{subfigure}{0.19\textwidth}
        \centering
        \includegraphics[width=\textwidth]{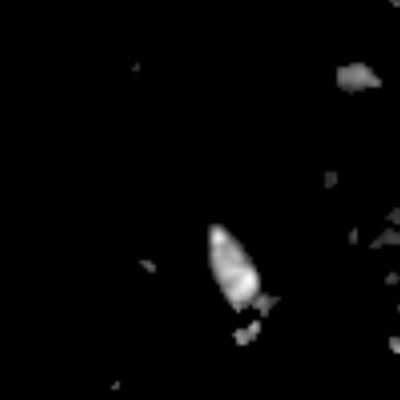}
        \caption{}
        \label{clust2}
    \end{subfigure}
    \begin{subfigure}{0.19\textwidth}
        \centering
        \includegraphics[width=\textwidth]{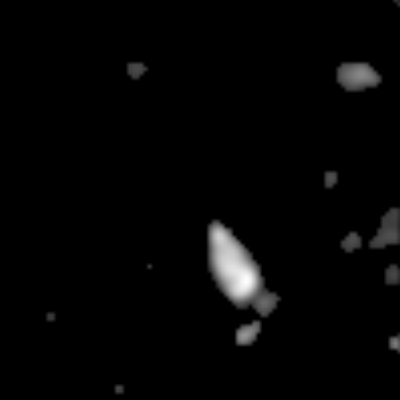}
        \caption{}
        \label{clust3}
    \end{subfigure}
    \begin{subfigure}{0.19\textwidth}
        \centering
        \includegraphics[width=\textwidth]{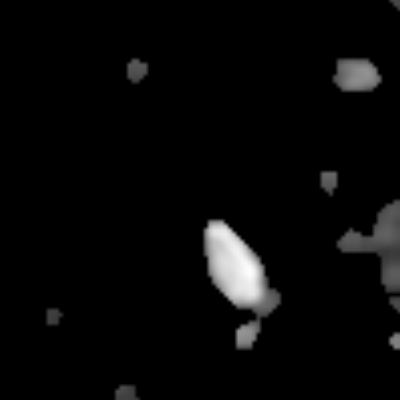}
        \caption{}
        \label{clust4}
    \end{subfigure}
    
    \caption{Progressive saliency maps by decoding consecutive blocks. Figures \ref{original} and \ref{truelabel} show the original and groundtruth images, respectively. Saliency maps obtained by decoding the output of blocks 1-4 for FLIM-BoFP in \ref{bofp1}-\ref{bofp4} and FLIM-Cluster in \ref{clust1}-\ref{clust4}.}
    \label{fig:prog-salie}
\end{figure}

This paper introduces FLIM-BoFP, a faster and more effective filter estimation method that replaces per-block clustering with a single clustering process at the input layer. It groups patches from marker pixels, storing cluster centers in a "Bag of Feature Points" (BoFP). Feature points are then mapped at each encoder block's input, where patches are extracted, and filters with biases are assigned per feature point, enabling precise placement and improved activation control. As shown in Figure \ref{fig:prog-salie}, FLIM-BoFP eliminates false positives more efficiently than \review{the per-block clustering approach, hereforth called  FLIM-Cluster}, enhancing object detection in saliency maps across consecutive blocks. 
\review{Previous work has shown that FLIM saliencies are suitable to initialize post-processing techniques, which take advantage of FLIM object detection capabilities and can improve initial saliency maps to fit the salient object's border better~\cite{FLIM-delineation,adaptive-decoders,salvagnini2025multilevelcellularautomataflim}}. Therefore, we also employ an object delineation algorithm that enhances post-processing for FLIM SOD models.

For evaluation, we consider a real-world problem that appears in the automated diagnosis of intestinal parasites -- the detection of parasite eggs and cysts among impurities. The experiments adopt as baselines SOD models from the state-of-the-art, including FLIM-Cluster, and three parasite datasets -- a challenging public dataset containing \textit{Schistosoma Mansoni} eggs\footnote{https://github.com/LIDS-Datasets/schistossoma-eggs}, and two private datasets with \textit{Entamoeba histolytica} cysts and \textit{Ancylostoma spp.} eggs. The private datasets are used to evaluate the generalization ability of the models pretrained in the public dataset. Schistosomiasis is a deadly neglected tropical disease caused by \textit{S. Mansoni}. Early diagnosis is crucial but typically relies on slow, manual stool examination~\cite{schistosomiase}. Automated microscopy produces thousands of images per slide, demanding highly efficient solutions. The proposed FLIM-BoFP-based SOD model addresses this need effectively.

 Section \ref{related-works} reviews existing literature on SOD and lightweight networks, including FLIM models. In Section \ref{flim-sod}, we introduce FLIM-Cluster SOD networks. Section \ref{flim-bofp} presents the FLIM-BoFP method. Sections \ref{experiments} and \ref{results} showcase our experimental setup and results. We state the conclusion and discuss future work in Section \ref{conclusion}.

\section{Related work}
\label{related-works}
Salient object detection (SOD) focuses on identifying visually prominent objects in images, aiding tasks like object recognition and scene analysis. Traditional methods, inspired by human vision, use local features like texture, color, and orientation in a bottom-up approach to generate saliency maps~\cite{survey-ullah}. 

Modern SOD uses convolutional encoders to extract multi-level, multi-scale features. Feature maps are combined via \review{Multilayer Perceptrons (MLPs)} for classification or convolutional decoders for saliency maps, with FCNs leading the latter. PiCANet employs a U-Net-like structure with a ResNet or VGG encoder~\cite{picanet}, BASNet refines boundaries using a ResNet-34 encoder and multi-layer supervision~\cite{qin2019basnet}, and U2Net enhances contextual information without pretrained backbones~\cite{u2net}.

Deep FCNs excel in accuracy but are computationally demanding, leading to lightweight SOD models that balance efficiency and performance. These models apply compression techniques \cite{chen2024review}, modify convolutions to reduce complexity, and leverage NAS (Neural Architecture Search) for \review{optimal architecture search} \cite{he2021automl}. Many leverage the MobileNet-V2 encoder, including SeaNet \cite{seanet} for remote sensing, HVPNet \cite{hvpnet} for real-time performance via hierarchical visual perception, and SAMNet \cite{samnet}, which enhances efficiency with stereoscopic multi-scale fusion, avoiding pretrained backbones.

Feature Learning from Image Markers (FLIM) assumes that effective filters can be estimated from markers placed on discriminative regions of a few representative images, such that designing lightweight networks from scratch with minimal data requirements should be possible~\cite{FLIM-og}. Furthermore, combining such an encoder with an adaptive decoder has shown effective for SOD, with flyweight networks considerably more efficient than existing lightweight models~\cite{adaptive-decoders}. Inspired by these results, the present study improves the effectiveness, efficiency, and explainability of filter estimation from image markers. 

\section{FLIM-Cluster SOD networks}
\label{flim-sod}

This section introduces the basic concepts of an FLIM-Cluster encoder and adaptive decoder used for the SOD models evaluated in this work. For the sake of space, and because this has already been formally explained in \cite{FLIM-og}, we will briefly summarize the key concepts. \review{We will start by introducing FLIM building blocks and how those concepts expand to define a convolutional encoder directly from an image, where extracted features may be adaptively decoded into saliency maps, eliminating the need for backpropagation.}

\subsection{Images, patches, and labeled markers}

\review{A 2D image $\hat{I} = (D_I, \mathbf{I})$ is represented as a grid of pixels, where each pixel $p=(x_p,y_p) \in D_I \subset \mathbb{Z}^2$ holds a feature vector with $m$ channels (e.g., RGB or activation channels), as $\mathbf{I}(p) = (I_1(p), I_2(p), \ldots, I_m(p)) \in \mathbb{R}^m$. Let $\mathcal{A}(p)$ express the adjacent set of a pixel $p$ -- a window centered at $p$ with $k\times k$ adjacent pixels $q$. 
\begin{equation}
\mathcal{A}(p) = \{q = (x_q, y_q) \in D_I \mid |x_q - x_p| \leq d \cdot \lfloor k/2 \rfloor, |y_q - y_p| \leq d \cdot \lfloor k/2 \rfloor\}
\end{equation}
where $k$ expresses the width and height of the neighborhood and $d$ is a dilation factor expressing the spacing between adjacent pixels. We can then define a patch $\mathbf{P}(p) \in \mathbb{R}^{k \times k \times m}$ as the concatenation of the feature vectors $\mathbf{I}(q) \forall q \in \mathcal{A}(p)$.}

 \review{The goal in the FLIM methodology is to learn convolutional filters directly from image patches extracted from marker pixels in discriminative regions. Given a set of representative training images $\mathcal{D}_T$, some pixels are manually annotated by "markers" $\mathcal{M}$ (labeled disks with a small radius). The positions of these markers guide filter learning in each convolutional block, and their labels (background or foreground) define the label of the filter used to create each channel, guiding the parameter estimation of adaptive decoders, which yields a saliency map.}

\subsection{FLIM-Cluster encoder}

The FLIM-Cluster encoder is a sequence of \textit{convolutional blocks}, each applying marker-based normalization, convolution with learned filters, activation (ReLU), and pooling (average or max-pooling).

\paragraph{\textbf{Marker-based normalization}}

\review{computes mean and standard deviation of each feature of marker pixels in $\mathcal{M}$. Those statistics are used to perform z-score normalization on input training images (first layer) or their feature maps (second layer onwards). The markers on input training images of the first block are mapped to the image domain of their feature maps at the input of the remaining blocks. Let $\mathcal{P}_C$ be the set of patches $\mathbf{P}(p)$ for all marker pixels $p\in \mathcal{M}$ using features at the input of a given block. Marker-based normalization is essentially a z-score normalization of those features in $\mathcal{P}_C$. Such an operation, using the statistics of the marker pixels, must be applied to normalize patches of any input image in that block, eliminating the need for bias computation in the block \cite{Joao2024UnderstandingMN}.}

\paragraph{\textbf{Convolution, activation, and pooling}}

\review{Let $\hat{I}$ be an image (or feature map) with $m$ channels after marker-based normalization in a given block. The kernels in the block must have the same shape $k\times k\times m$. Hence, the convolution between image $\hat{I}=(D_I,\mathbf{I})$ and each kernel $\mathbf{K}_i\in \mathbb{R}^{k\times k\times m}$, $i=1,2,\ldots,m'$, of a kernel bank with $m'$ kernels defines a feature map $\hat{J}=(D_I,\mathbf{J})$, where $\mathbf{J}(p)=(J_1(p).J_2(p),\ldots,J_{m'}(p))$ and 
\begin{eqnarray}
J_i(p)&=&\langle \mathbf{P}(p),  \mathbf{K}_i \rangle,
\end{eqnarray}
where  $\mathbf{P}(p)$ is the normalized patch at position $p\in D_I$.}

 \review{Such kernels $\mathbf{K}_i$ are determined from the training images in $\mathcal{D}_T$ as follows. Let $\mathcal{M}_I$ be a marker (disk) in a training image $\hat{I}$ (or feature map) after marker-based normalization in a given block. Let  $\mathcal{P}_I$ be the set of normalized patches $\mathbf{P}(p)\in \mathbb{R}^{k\times k\times m}$ extracted from marker pixels $p\in \mathcal{M}_I$. FLIM use the K-means algorithm to find $c$ groups of patches in $\mathcal{P}_I$. The center of each group is forced to have a unit norm and used as a kernel $\mathbf{K}_i$. By that, each marker generates $c$ kernels. The procedure is repeated by marker to obtain the total number $m'$ of kernels in the block. The number $c$ of kernels per marker is a hyperparameter of the model, together with $k$, $d$, pooling type, adjacency, and stride, but the number $m'$ of kernels in a block will depend on the number of markers drawn by the user. In this work, we fix the same number $m'$ of kernels in each block.}

\review{Learning convolutional filters directly from image patches exploits the geometric properties of convolution operations to identify discriminative visual patterns. From a geometric perspective, each kernel $\mathbf{K}_i$ is orthogonal to a hyperplane positioned at the origin of $\mathbb{R}^{k\times k \times m}$. Simultaneously, normalized image patches $\mathbf{P}(p)$ correspond to points distributed around the origin of this local feature space. The inner product $\langle P(p), \mathbf{K}_i \rangle$ determines the signed distance between each normalized patch and the hyperplane associated with kernel $\mathbf{K}_i$. The resulting feature values $J_i(p)$ are submitted to ReLU activation, eliminating patches on the negative side of the hyperplane. Patches exhibiting visual patterns similar to the learned kernel reside on the positive side of the hyperplane. By that, image regions with patterns of interest should be activated. The choice of kernels from cluster centers at each marker makes them representative visual patterns of interest. After each channel $J_i(p)$ is submitted to ReLU activation, max/average pooling is applied to the resulting feature map, aggregating nearby activations.}

A filter's label, derived from its corresponding marker, determines whether its activation channel highlights the foreground or background in the feature map. Foreground filters brighten objects, while background kernels do the opposite, though separation may be imperfect due to image variability. Adaptive decoders address this issue using a heuristic function to dynamically identify which channels in a feature map can be classified as foreground, background, or undefined channels.

\subsection{Adaptive decoder}\label{sec:adpdecoder}

An adaptive decoder may apply point-wise convolution (weighted average) to combine feature map channels, followed by ReLU activation. Some estimate pixel-wise weights per channel before averaging. This work adopts an example of the latter approach, the mean-based decoder from \cite{adaptive-decoders}.

\review{The mean-based decoder processes a feature map with $m'$ channels by computing mean activations,  $\mu_F(p)$ and  $\mu_B(p)$, for foreground and background channels within a small neighborhood of each pixel $p$. The method assigns positive weights $\alpha_i(p)$ to pixels in foreground channels when \( \mu_F(p) > \mu_B(p) \), as it is expected that object pixels satisfy this condition in foreground channels. Conversely, it assigns negative weights $\alpha_i(p)$ to pixels in background channels when \( \mu_F(p) < \mu_B(p) \), as it is expected that background pixels satisfy this condition in background channels. Otherwise, in the absence of evidence about the pixel's label, the weight of the pixel is assigned to zero. Averaging channels with pixel-wise weights and applying ReLU tends to isolate true positives in the saliency map (Figure~\ref{fig:prog-salie}). In Equation \ref{eq:decoder}~\cite{adaptive-decoders}, channel $J_i$ belongs to the output feature map $\hat{J}$ of the last block and $\lambda(J_i(p))\in \{1,2\}$ is the label of kernel $\mathbf{K}_i$, where $1$ is foreground and $2$ is background, as derived from the corresponding labeled marker. }

\review{\begin{eqnarray}
\label{eq:decoder}
\alpha_i(p) & = & \begin{cases}
+1, & \mbox{if $\lambda(J_i(p)) = 1$ \& $\mu_F(p) > \mu_B(p)$},\\
-1, & \mbox{if $\lambda(J_i(p)) = 2$ \& $\mu_F(p) < \mu_B(p)$},\\
0, & \mbox{otherwise.}
\end{cases}
\end{eqnarray}}

\section{FLIM encoders with Bag of Feature Points}

\label{flim-bofp}

\review{The training of a FLIM encoder with BoFP consists of two steps: (i) clustering-based feature point estimation from each marker to identify discriminative locations in each training image $\hat{I} \in \mathcal{D}_T$, storing the feature points of all markers in the image into a set $\mathcal{B}_I$, and (ii) direct filter estimation from mapped feature points at each block's input. The complete BoFP is the union set $\mathcal{B}$.}

\review{\begin{equation}
\mathcal{B} = \bigcup_{I \in \mathcal{D}_T} \mathcal{B}_I 
\end{equation}}

\begin{figure}
 \begin{center}
\includegraphics[width=0.7\textwidth]{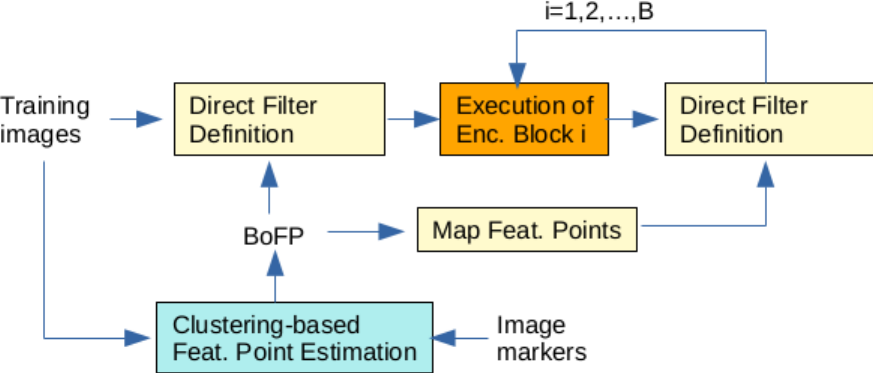}
\end{center}
\caption{Training of a FLIM encoder with BoFP begins with a single clustering-based feature point estimation to insert discriminative image locations in the BoFP. Filters for each block are then directly derived from mapped points in the feature map at the block’s input, ensuring precise filter placement and efficient processing.}
\label{fig:bofptraining}
\end{figure}

\review{Figure~\ref{fig:bofptraining} shows each step in different colors. Step (i) runs once, while step (ii) repeats for each block $i$, using the previous block's output as input. The key difference from FLIM-Cluster lies in the control for filter estimation. As set $\mathcal{B}$ is defined only once, all layers derive convolutional filters from the same spatial locations mapped across blocks. This choice is based on the assumption that if a patch $\mathbf{P}(p)$ at location $p$ of the input image defines a discriminative filter, the spatial location $p$ remains important in subsequent feature maps.}

\review{In step (i), for a given number $c$ of kernels per marker, patches are extracted from marker pixels and grouped into $c$ clusters by the $K$-means algorithm, but with no need for marker-based normalization. Each cluster center identifies a discriminative location $p$ in a training image $\hat{I}\in \mathcal{D}_T$, whose patch $\mathbf{P}(p)$ is the closest to that center. The location $p$ is then inserted into the set $\mathcal{B}_I$.}

\review{In step (ii), the points in $\mathcal{B}_I$ are mapped to the input of each block for kernel estimation. Let $\hat{I}=(D_I,\mathbf{I})$ be a training image with $m$ channels (or a feature map) at the input of a given block. From the feature points $p$ in $\mathcal{B}_I$, FLIM-BoFP extract patches $\mathbf{P}(p)\in \mathbb{R}^{k \times k\times m}$ to compose a patch set $\mathcal{P}_B$. Let $\mathbf{\mu}_B$ and $\mathbf{\sigma}_B$ be the mean and standard deviation of the features in $\mathcal{P}_B$. They are used for z-score normalization of those patches. Afterwards, each normalized patch  $\mathbf{P}(p)$ generates a kernel $\mathbf{K}_i$ with bias $b_i$ as follows. 
\begin{eqnarray}
\mathbf{K}_{i} & = & \frac{\mathbf{P}(p)}{\|\mathbf{P}(p)\|} \oslash \mathbf{\sigma}_B, \\
b_i & = & -\langle \mathbf{\mu}_B, \mathbf{K}_i \rangle,    
\end{eqnarray}
where $\oslash$ is the element-wise division and $\langle \cdot \rangle$ is the inner product. The kernel bank with $m'$ kernels in the block is the union of all kernels $\mathbf{K}_{i}$ from all training images $\hat{I}\in \mathcal{D}_T$, using all $m'$ points in $\mathcal{B}$. Again, each block has a fixed number $m'$ of kernels. This procedure eliminates the need for marker-based normalization in each block, since it uses the statistics derived from the patches in $\mathcal{P}_B$ to create kernels and biases. After convolution with the kernel bank, $J_i(p) + b_i$ is submitted to ReLU activation, and pooling is applied to obtain the output feature map of the block.} 

\section{Experimental Setup}
\label{experiments}

This section describes the experimental setup to evaluate FLIM-BoFP for parasite detection in optical microscopy images. FLIM SOD networks consist of an encoder trained with FLIM-BoFP or FLIM-Cluster followed by the mean-based adaptive decoder. The encoders contain almost the same operations, except that FLIM-BoFP dismisses marker-based normalization per block since it estimates bias for each filter. Markers are usually strokes or disks. In this work, we use disks with a small radius of 3.0 pixels for both methods to improve control over filter location and, consequently, explainability. 

\paragraph{Parasite datasets}

\reviewData{
The experiments involve three datasets of  parasites with food debris (impurities), such that the first dataset is used for training and testing, and the others are used to evaluate the generalization ability of the models.
\begin{enumerate}
    \item \textit{S. Mansoni}, a challenging dataset, used for training and evaluation, with 1,219 RGB images ($400\times 400$ pixels) containing none, one, or multiple eggs of \textit{Schistosoma Mansoni}~\cite{smansoni}. 
    \item \textit{Entamoeba}, a private dataset with 395 RGB images ($256\times 256$ pixels) containing a single \textit{Entamoeba Hystolitica} cyst per image. 
    \item \textit{Ancylostoma}, a private dataset with 320 RGB images ($400\times 400$ pixels) containing one or multiple eggs of \textit{Ancylostoma} spp. per image.
\end{enumerate}     
}
\review{The \textit{S. Mansoni} dataset was randomly split into training and testing sets, with each comprising 50\% of the data ($\mathcal{X}_1$ and $\mathcal{X}_2$, respectively). The training set $\mathcal{X}_1$ served multiple purposes: (i) a few representative images in $\mathcal{X}_1$ were selected for the set $\mathcal{D}_T\subset \mathcal{X}_1$ used to train and find the best architectures for FLIM-BoFP and FLIM-Cluster, with different images chosen across experimental runs to ensure robust marker-based training; (ii) set $\mathcal{D}_T\subset \mathcal{X}_1$ is the same used to train/fine-tune the deep-learning models; (iii) the remaining images in $\mathcal{X}_1\setminus \mathcal{D}_T$ served as validation set for both FLIM-based and deep-learning models. The testing set $\mathcal{X}_2$ served for the final performance evaluation of all methods. The remaining datasets were used to evaluate whether models pretrained on \textit{S. Mansoni} can generalize to a different parasite detection task through zero-shot evaluation, representing a transfer learning scenario within the same parasitology application domain.}

\paragraph{Training image selection}

\review{The algorithm in~\cite{decoder-1} was used to iteratively select training examples from $\mathcal{X}_1$ (\textit{S. Mansoni}), refining the model with each addition while keeping only images that improved performance on the validation set. We start by adding one randomly selected image from $\mathcal{X}_1$ to $\mathcal{D}_T$, then we assess the validation metrics on the images from $\mathcal{X}_1 \setminus \mathcal{D}_T$. Among the images for which the method obtained low performance, we select an additional image to compose $\mathcal{D}_T$. This procedure is repeated until the user verifies satisfactory performance. FLIM-Cluster was trained with 3-4 images per split, and FLIM-BoFP with 2-3 images per split.}

\paragraph{Compared Methods}

FLIM-BoFP-based SOD networks were compared with FLIM-Cluster and deep-learning models: HVPNet~\cite{hvpnet}, SAMNet~\cite{samnet}, SeaNet~\cite{seanet}, and U2-Net~\cite{u2net}, known for strong performance with small training sets. FLIM models were trained from scratch, while the others used recommended hyperparameters, boundary-aware loss, and pretrained backbones (when applicable). Fine-tuning of the deep-learning models utilized the combined representative training images from FLIM-BoFP and FLIM-Cluster, providing 5-6 images in $\mathcal{D}_T$ per split for training models for S. Mansoni. To improve SAMNet and HVPNet's performance, training was extended to 400 epochs, while SeaNet remained with 50 epochs due to no observed benefit from longer training. U2-Net was trained for 1000 epochs, and the model with the minimum loss value on validation was selected. \review{Lastly, we chose not to evaluate the state-of-the-art iterative segmentation methods, as SAM~\cite{sam} and MedSAM~\cite{medsam}, since those models are not suitable for execution under limited resource scenarios (\eg low-cost computers commonly found in clinical analysis laboratories of developing countries).}

\paragraph{Post-processing}

\review{FLIM SOD networks benefit from post-processing algorithms that enhance object boundary delineation, as they do not employ boundary-aware losses. We use Dynamic Trees (DT) ~\cite{dynamic-trees} as a competitive solution for fair comparison to baseline models. In DT, saliency maps serve as seed initialization within an optimum-path forest framework. Root pixels are extracted through Otsu thresholding, morphological operations (erosion and dilation), and area filtering --- components within the range [1000, 9000] --- define internal seeds, while their complement forms external seeds. Such area filters were also used to reduce false positives from the deep-learning models.}
 
\paragraph{\textbf{Performance Metrics}} 

To compare the models, we measure their performance using popular metrics such as F-score, Mean Absolute Error (MAE), and weighted F-score (wF). Due to the unique properties of the dataset, namely the typically off-center object locations, some commonly used metrics, such as E-measure, which rely on centered objects, were not considered. Otsu thresholding was used to binarize the saliency maps necessary to calculate the F-score. Metrics were calculated using the three-split average for the \textit{S. Mansoni} dataset for all models. \review{For \textit{Entamoeba} and \textit{Ancylostoma}, we apply the three best split's architectures found for \textit{S. Mansoni} and average their results to present the metrics of the FLIM models.}

\section{Results and discussion}
\label{results}
\captionsetup[subfigure]{skip=0pt}
\begin{figure}[!h]
    \centering
    \begin{subfigure}{0.45\textwidth}
        \centering
         \includegraphics[width=\textwidth]{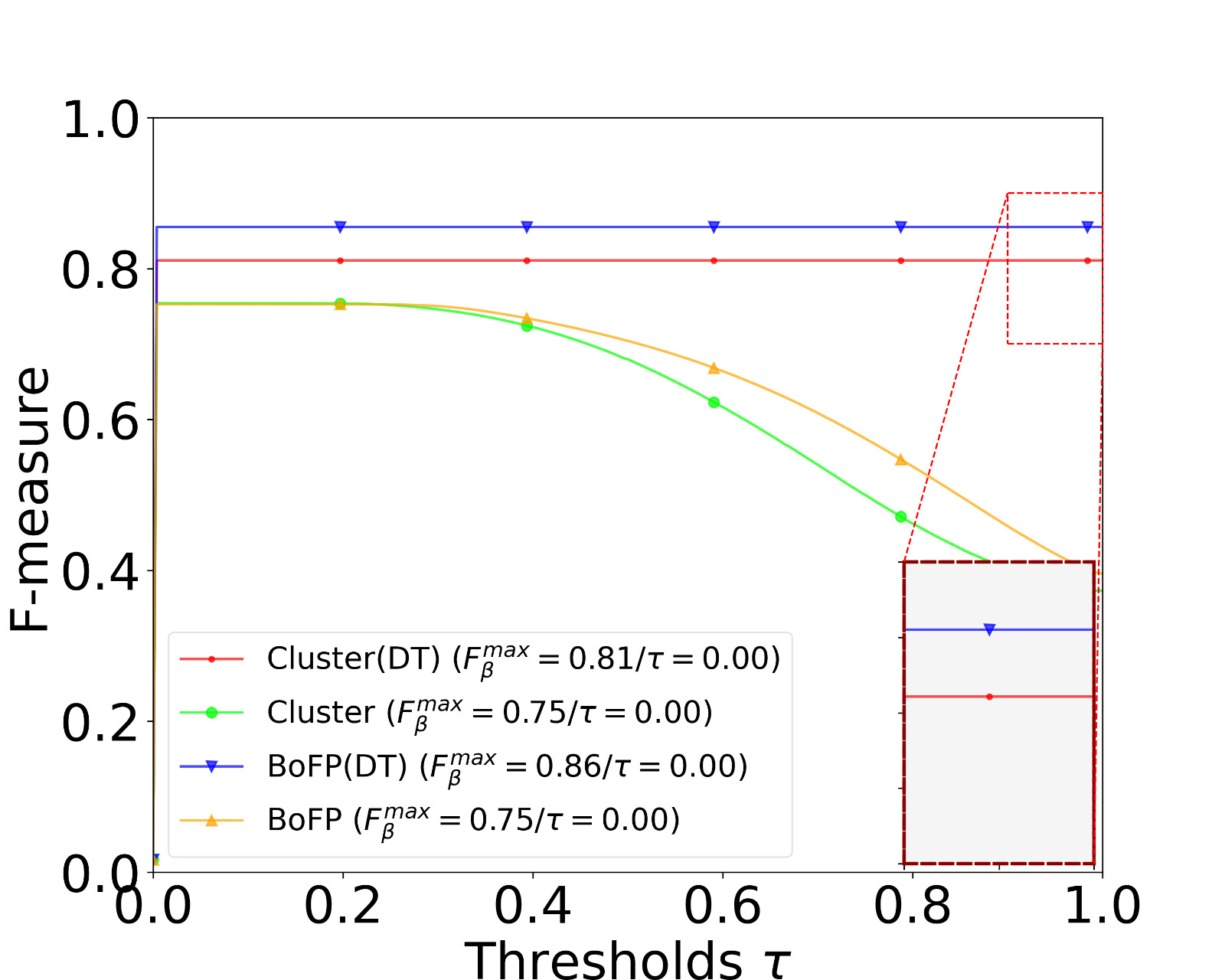}
        \caption{}
        \label{fig:before_after_post}
    \end{subfigure}
    \begin{subfigure}{0.45\textwidth}
        \centering
        \includegraphics[width=\textwidth]{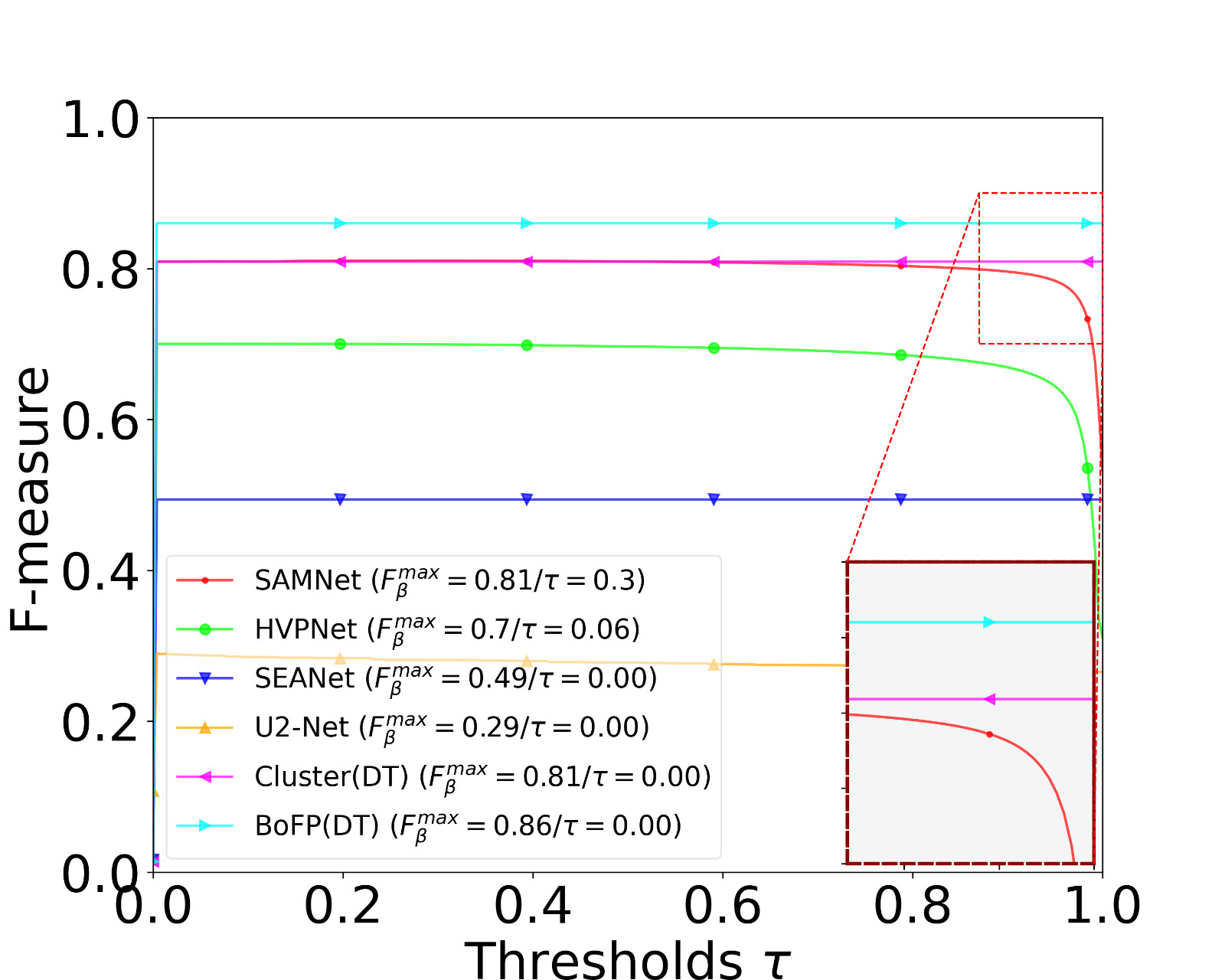}
        \caption{}
        \label{fig:compare_dt}
    \end{subfigure}
    \caption{Average F-measure curves across splits on test sets for each model. Figure \ref{fig:before_after_post} compares FLIM models with and without DT post-processing for \textit{S. Mansoni}. Figure \ref{fig:compare_dt} contrasts FLIM$_{DT}$ models against deep-learning models on \textit{S. Mansoni}, with the FLIM models at the top. 
    }

    \label{fig:ablation_graphs}
\end{figure}

This section presents an ablation study to determine the optimal architecture for each FLIM model per split and analyze the impact of DT. It includes both quantitative and qualitative results, along with a comparative evaluation of the models.

\subsection{Ablation study}

For FLIM models using $\mathcal{X}_1$ (\textit{S. Mansoni}), a grid search was conducted across key hyperparameters -- kernel size, number of kernels per marker, and number of blocks -- to optimize architectures. Kernel sizes between $3\times 3$ and $7\times 7$ were tested, along with 1 to 4 kernels per marker and up to 4 blocks per split. The best architectures in each split were used for evaluation, with $3\times 3$ emerging as the optimal kernel size across all models. FLIM-Cluster performed best with 1 or 3 kernels per marker and up to 2 convolutional blocks. In comparison, FLIM-BoFP used 1 or 2 kernels per marker with 2 or 4 convolutional blocks, making it slightly heavier due to its deeper architecture. \reviewData{The same architectures were applied to the three datasets.} The specific architectures for each encoder and split can be seen in our GitHub (released after review).

The F-measure (F-score with $\beta=1$) was computed after post-processing in $P_1$ to select the best FLIM model for each split, identifying BoFP1-BoFP3 and Cluster1-Cluster3 (with DT) as the top performers. Average curves of F-measure (F-score with $\beta=0.3$) across 255 thresholds on the saliency maps of the test sets show the DT's positive impact on model robustness (Figure~\ref{fig:before_after_post}). A significant variation in optimal binarization thresholds suggests that the adaptive decoder enhances saliency distribution across the object of interest. The curves show a superior performance of the FLIM models with DT under limited-training-data conditions compared to the deep-learning networks in both datasets, with the FLIM-BoFP models at the top in performance \reviewData{(Figure~\ref{fig:compare_dt})}. 

\subsection{Comparative analysis}

\begin{figure}[!htb]
\begin{center}
\newcommand\sizefig{0.121\textwidth}
\centering

\renewcommand{\arraystretch}{1}
\begin{tabular}{*{8}{c@{\hskip 0.5pt}}c}
\includegraphics[width=\sizefig]{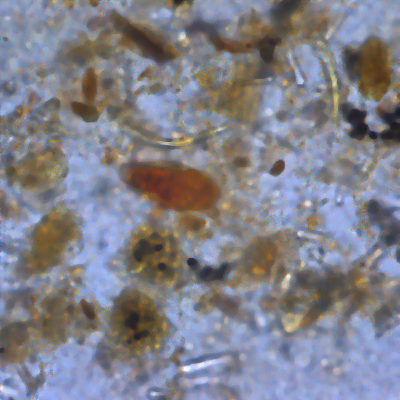}&
\includegraphics[width=\sizefig]{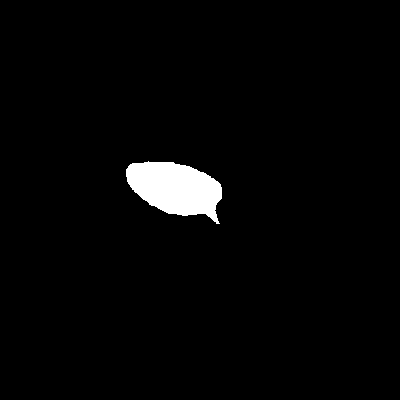}&
\includegraphics[width=\sizefig]{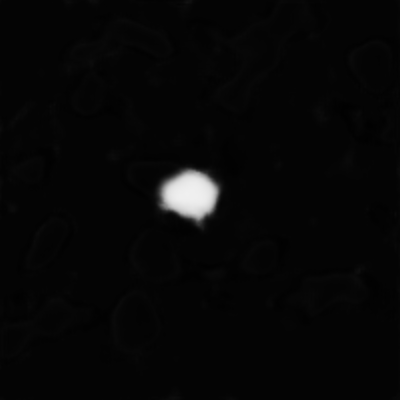}&
\includegraphics[width=\sizefig]{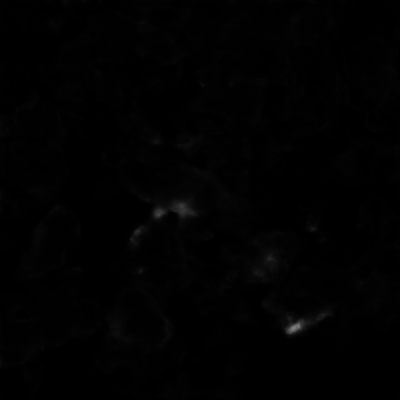}&
\includegraphics[width=\sizefig]{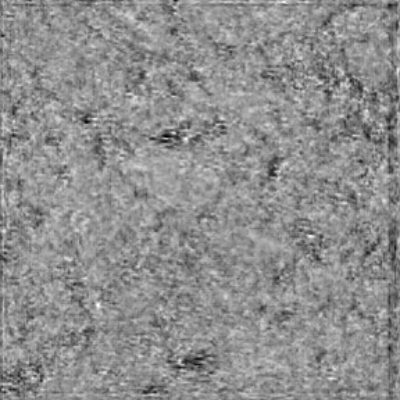}&
\includegraphics[width=\sizefig]{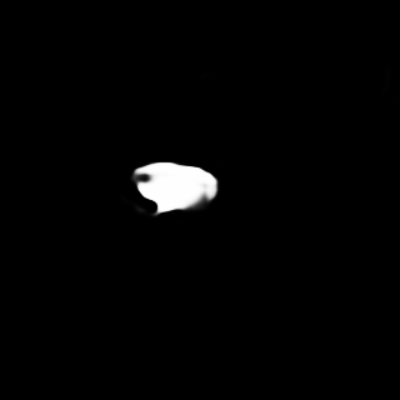}&
\includegraphics[width=\sizefig]{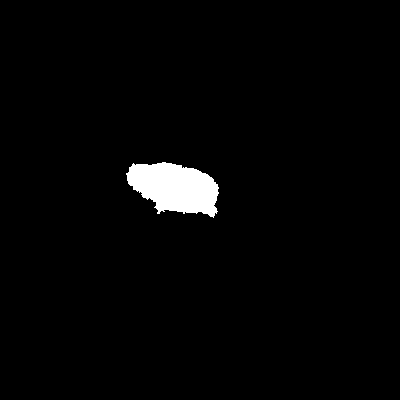}&
\includegraphics[width=\sizefig]{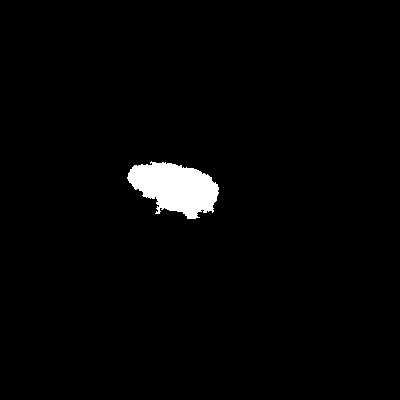}\\
\includegraphics[width=\sizefig]{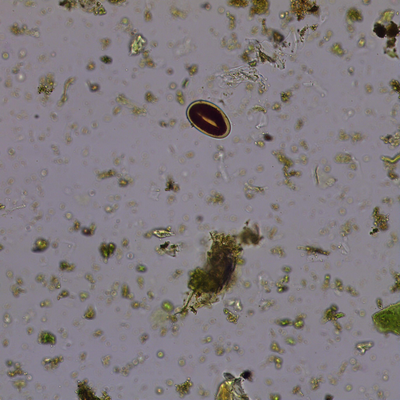}&
\includegraphics[width=\sizefig]{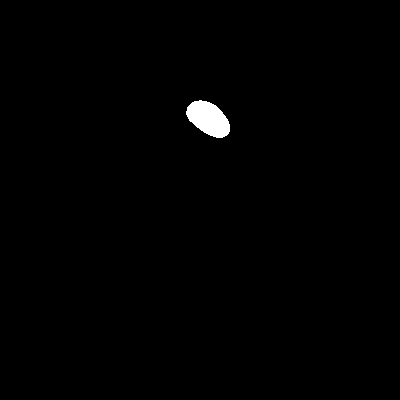}&
\includegraphics[width=\sizefig]{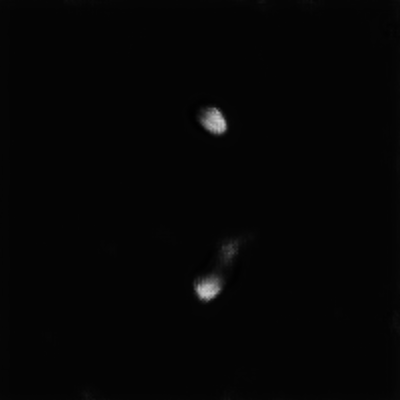}&
\includegraphics[width=\sizefig]{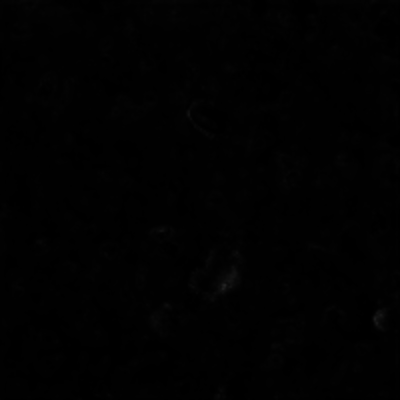}&
\includegraphics[width=\sizefig]{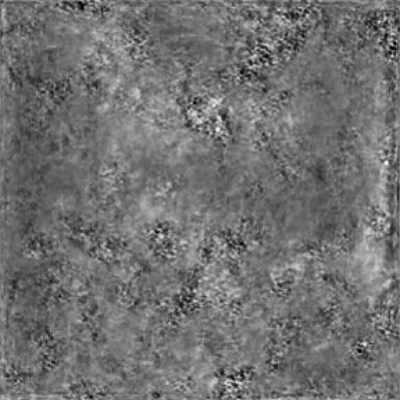}&
\includegraphics[width=\sizefig]{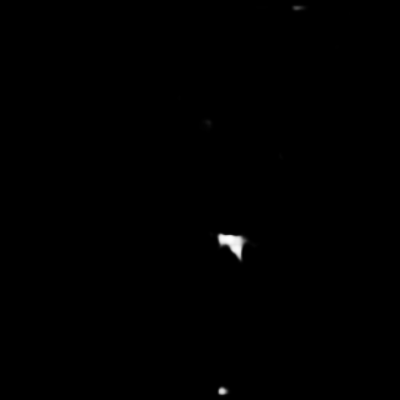}&
\includegraphics[width=\sizefig]{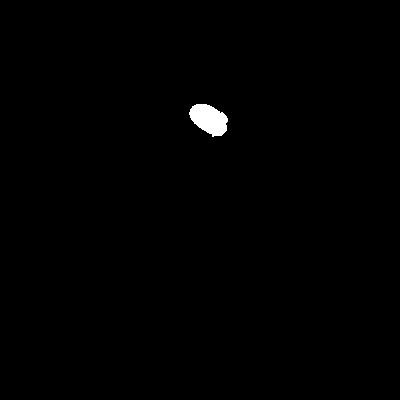}&
\includegraphics[width=\sizefig]{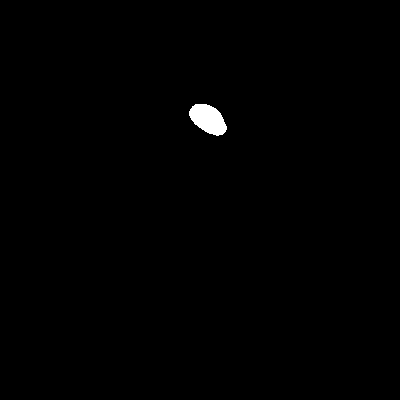}\\
\includegraphics[width=\sizefig]{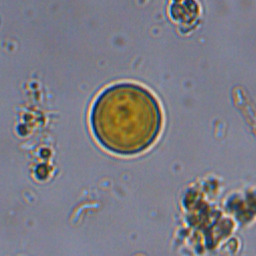}&
\includegraphics[width=\sizefig]{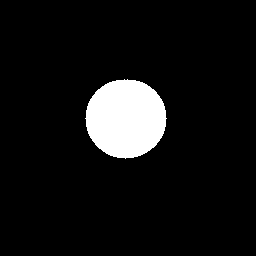}&
\includegraphics[width=\sizefig]{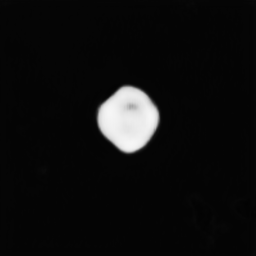}&
\includegraphics[width=\sizefig]{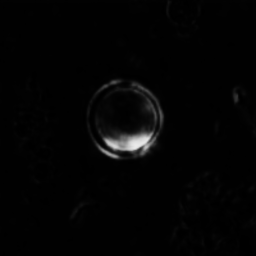}&
\includegraphics[width=\sizefig]{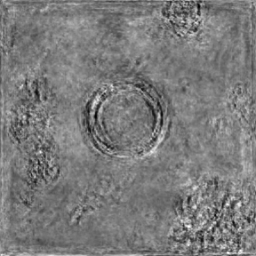}&
\includegraphics[width=\sizefig]{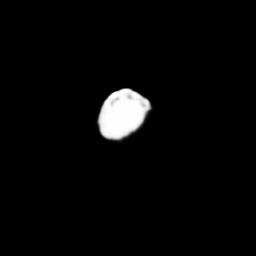}&
\includegraphics[width=\sizefig]{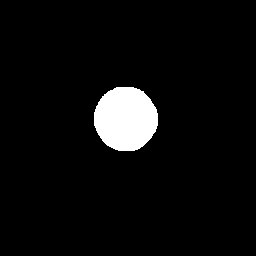}&
\includegraphics[width=\sizefig]{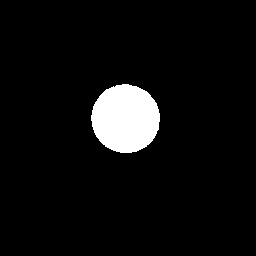}&
\\
(a) Image & \footnotesize (b) GT & SAMNet & HPVNet & SEANet & U2Net &Cluster &BoFP\\
\end{tabular}

\caption{\reviewData{Qualitative detection results on test images of the \textit{S. Mansoni}, \textit{Entamoeba} and \textit{Ancylostoma} datasets. \textit{GT} is the ground-truth segmentation mask.}}
    \label{fig:qualitative}
\end{center}
\end{figure}

The results in Table \ref{tab:flim-baselines-schisto} reveal a clear contrast in model sizes, with FLIM models significantly smaller than the lightest deep-learning models. Despite more training images and ground-truth-mask-based fine-tuning, deep-learning models were outperformed, with FLIM-BoFP consistently achieving the best metrics (1st or 2nd place). Overfitting likely impacted deep-learning models due to limited training data—SEANet, for instance, generated random noise for images outside its training set. After applying the area filter, SEANet's saliency maps were entirely nullified (Figure \ref{fig:qualitative}), explaining its poor performance across datasets. DT's effect on FLIM models is evident in larger improvements on average metrics across all splits.

\reviewData{For the \textit{Entamoeba} and \textit{Ancylostoma} datasets, experiments highlight the higher resilience of the FLIM models, particularly FLIM-BoFP, when generalizing to new datasets of the same application domain (Table \ref{tab:flim-baselines-entamoeba}). The 0-shot experiment shows FLIM-based models were more resilient than deep-learning models —SAMNet, a top performer in \textit{S. Mansoni} performed poorly. HVPNet maintained performance better, but still lagged behind FLIM-BoFP and -Cluster. FLIM-BoFP remained the best-performing model in all metrics. Figure \ref{fig:qualitative} demonstrates the qualitative performance and shows the superior capacity of FLIM to generalize to unseen data, in the microscopy domain, even when related to other parasites.}

\begin{table}[!h]
\caption{The average results from three \textit{S. mansoni} test sets (one per split) for F-score (F$_\beta$, $\beta=0.3$), Mean Absolute Error (MAE), and Weighted F-measure (wF). Performance is color-coded: best in \green{green}, second-best in \blue{blue}, and worst in \red{red}. Arrows $\uparrow$ and $\downarrow$ indicate whether higher or lower values are preferable. The "M" and "K" notations represent millions and thousands of trainable parameters, respectively, with FLIM models showing the average number across splits of trainable parameters per model.}
\centering
\vspace{1em}
\begin{adjustbox}{width=0.85\columnwidth}
\rowcolors{2}{gray!20}{white}
\begin{tabularx}{\columnwidth}{>{\hsize=2\hsize}X | c | *{3}{c}}
    \hline \hline
    \hiderowcolors 
    \multicolumn{1}{X|}{Model} & \multicolumn{1}{c|}{\# Parameters} & \multicolumn{3}{c}{\textit{S. mansoni}} \\ \cline{3-5}
        &  & F$_\beta$$\uparrow$ & MAE$\downarrow$ & wF$\uparrow$ \\ \hline
    \showrowcolors
    SAMNet                  & 1.33 M & 0.824$\pm$0.120   & \blue{0.006$\pm$0.000}   & 0.786$\pm$0.120 \\
    HVPNet                  & 1.23 M & 0.710$\pm$0.041          & \green{0.005$\pm$0.000}    & 0.668$\pm$0.049 \\
    SEANet                  & 2.76 M & 0.494$\pm$0.250          & 0.013$\pm$0.000           & 0.494$\pm$0.250 \\
    U2-Net                  & 44 M   & 0.658$\pm$0.151          & 0.013$\pm$0.005           & 0.623$\pm$0.148 \\
    FLIM-Cluster            & 5.677$\pm$5.720 K    & \red{0.368$\pm$0.155} & \red{0.014$\pm$0.000} & \red{0.198$\pm$0.079} \\
    FLIM-Cluster$_{DT}$ & 5.677$\pm$5.720 K    & \blue{0.844$\pm$0.017}    & \blue{0.006$\pm$0.000}           & \blue{0.837$\pm$0.015} \\
    FLIM-BoFP               &  31.345  $\pm$  7.888 K    & 0.497$\pm$0.190 & 0.012$\pm$0.000 & 0.327$\pm$0.128 \\
    FLIM-BoFP$_{DT}$    &  31.345  $\pm$  7.888 K     & \green{0.860$\pm$0.091}          & \blue{0.006$\pm$0.000}    & \green{0.847$\pm$0.089} \\
    \hline \hline
\end{tabularx}
\end{adjustbox} 
\label{tab:flim-baselines-schisto}
\end{table}

\begin{table}[!h]
\caption{
 The results in \textit{Entamoeba} and \textit{Ancylostoma} as  test sets, using none training image. For the FLIM models, these are the average results of the best architectures of each split, as found in \textit{S. Mansoni}. }
\centering
\vspace{1em}
\begin{adjustbox}{width=1\textwidth}
\rowcolors{2}{gray!20}{white}
\begin{tabularx}{1.1\textwidth}{l|*{6}{c}}
    \hline \hline
    \hiderowcolors 
    \multicolumn{1}{l|}{Model} & \multicolumn{3}{c|}{\textit{Entamoeba}} & \multicolumn{3}{c|}{\textit{Ancylostoma}} \\ \cline{2-7}
     & F$_\beta$$\uparrow$ & MAE$\downarrow$ & wF$\uparrow$ & F$_\beta$$\uparrow$ & MAE$\downarrow$ & wF$\uparrow$ \\ \hline
    \showrowcolors
    SAMNet                  & 0.331$\pm$0.176           & 0.073$\pm$0.001         & 0.212$\pm$0.094
    & 0.085$\pm$0.079           & 0.022$\pm$0.002         & 0.018$\pm$0.015\\
    HVPNet                  & \blue{0.682$\pm$0.066}           & 0.080$\pm$0.002         & 0.256$\pm$0.050 & 0.286$\pm$0.155&0.012$\pm$0.004&0.152$\pm$0.093\\
    SEANet                  & \red{0.000$\pm$0.000}     & \red{0.086$\pm$0.001}   & \red{0.000$\pm$0.000}
    & \red{0.007$\pm$0.000}     & \red{0.506$\pm$0.038}   & \red{0.007$\pm$0.000}\\
    U2-Net                  & 0.362$\pm$0.098           & 0.077$\pm$0.004         & 0.210$\pm$0.060& 0.082$\pm$0.047           & \blue{0.013$\pm$0.008}         & 0.036$\pm$0.016\\
    FLIM-Cluster            &  0.596$\pm$0.093           & 0.078$\pm$0.003         & 0.215$\pm$0.055 & 0.362$\pm$0.073 & \green{0.012$\pm$0.002} & 0.279$\pm$0.031\\
    FLIM-Cluster$_{DT}$            &  0.604$\pm$0.218           & \blue{0.046$\pm$0.001}         & \blue{0.542$\pm$0.177} & \blue{0.456$\pm$0.036} & 0.069$\pm$0.000 & \blue{0.437$\pm$0.036}\\
    FLIM-BoFP               & 0.657$\pm$0.046 & 0.071$\pm$0.002 & 0.278$\pm$0.043  & 0.345$\pm$0.023 & 0.014$\pm$0.001 & 0.274$\pm$0.014 \\
    FLIM-BoFP$_{DT}$               & \green{0.792$\pm$0.125}    & \green{0.036$\pm$0.001} & \green{0.697$\pm$0.105}  & \green{0.517$\pm$0.010} & 0.069$\pm$0.000 & \green{0.492$\pm$0.021} \\
    \hline \hline
\end{tabularx}
\end{adjustbox} 
\label{tab:flim-baselines-entamoeba}
\end{table}

\section{Conclusion}
\label{conclusion}

We introduced FLIM-BoFP, a novel kernel estimation technique for FLIM methodology, designed as a more efficient alternative to FLIM-Cluster. FLIM-BoFP demonstrated superior efficiency and effectiveness, outperforming FLIM-Cluster and several heavier deep-learning models. Despite having significantly fewer parameters, it excelled in salient object detection tasks involving microscopy images of parasite eggs and cysts. On the public \textit{S. Mansoni} dataset, FLIM-BoFP surpassed the best deep-learning model, using less than 3\% of the trainable parameters. Additionally, FLIM models demonstrated to be more resilient than deep-learning models, when generalizing to datasets of the same application domain,\textit{Entamoeba} and \textit{Ancylostoma}. 

We may conclude that FLIM models, particularly the top performer, FLIM-BoFP, are valuable for medical object detection in data- and resource-limited environments.
Future work will explore integrating shape-prior morphometric filters, leveraging the combination of multiple-block decoding to enhance saliency maps, and evaluating model generalization on more datasets of parasites. We also intend to examine how user-based image marking impacts performance. FLIM-BoFP’s efficiency also makes it promising for pseudo-labeling large datasets to aid in training heavyweight models and optimizing encoder architectures before training.


\begin{credits}
\subsubsection{\ackname} The authors acknowledge grants from FAPESP (2023/14427-8, 2024/08332-7, and 2024/23772-3), CAPES (88887.820891/2023-00), and CNPq (304711\slash2023-3). The authors also acknowledge grants from the Eldorado Research Institute.

\subsubsection{\discintname}
The authors have no competing interests to declare that are relevant to the content of this article.
\end{credits}
%
%
%
\bibliographystyle{splncs04}
\bibliography{mybibliography}
\end{document}